\documentclass[runningheads,a4paper]{llncs}

\usepackage{amssymb}
\usepackage{graphicx}
\usepackage{wrapfig}
\usepackage[tight]{subfigure}
\usepackage{microtype}
\usepackage{algorithm}
\usepackage{algorithmic}
\usepackage{enumerate}
\usepackage{amsmath}
\usepackage{amsfonts}
\usepackage[font=small,labelfont=bf]{caption}
\usepackage[usenames,dvipsnames]{color}
\usepackage[english]{babel}
\usepackage{wrapfig}

\def\ie{{i.e.}}
\def\eg{{e.g.}~}

\graphicspath{{f/}}

\newcommand{\myparagraph}[1]{\smallskip\noindent\textbf{#1.} }

\begin{document}
\mainmatter  

\title{Increased HARDI Acceleration by ($k,q$)-CS with a Joint Spatial-Angular Transform}
\title{($k,q$)-Compressed Sensing for HARDI using a Joint Spatial-Angular Analysis-Synthesis Model}
\title{($k,q$)-Compressed Sensing for HARDI using a Joint Spatial-Angular Representation}
\title{($k,q$)-Compressed Sensing for dMRI with Joint Spatial-Angular Sparsity Prior}


\titlerunning{($k,q$)-CS with Joint Spatial-Angular Sparsity}
%
%
\author{Evan Schwab, Ren\'{e} Vidal, Nicolas Charon}

%
\institute{Center for Imaging Science, Johns Hopkins University}
%
%
\toctitle{Lecture Notes in Computer Science}
\tocauthor{Authors' Instructions}

\maketitle

\section*{Abstract}

Advanced diffusion magnetic resonance imaging (dMRI) techniques, like diffusion spectrum imaging (DSI) and high angular resolution diffusion imaging (HARDI), remain underutilized compared to diffusion tensor imaging because the scan times needed to produce accurate estimations of fiber orientation are significantly longer. To accelerate DSI and HARDI, recent methods from compressed sensing (CS) exploit a sparse underlying representation of the data in the spatial and angular domains to undersample in the respective $k$- and $q$-spaces. State-of-the-art frameworks, however, impose sparsity in the spatial and angular domains \textit{separately} and involve the sum of the corresponding sparse regularizers. In contrast, we propose a unified ($k,q$)-CS formulation which imposes sparsity \textit{jointly} in the spatial-angular domain to further increase sparsity of dMRI signals and reduce the required subsampling rate. To efficiently solve this large-scale global reconstruction problem, we introduce a novel adaptation of the FISTA algorithm that exploits dictionary separability. We show on phantom and real HARDI data that our approach achieves significantly more accurate signal reconstructions than the state of the art while sampling only $2$-$4\%$ of the ($k,q$)-space, allowing for the potential of new levels of dMRI acceleration\index{acceleration}. 

\section{Introduction}
Diffusion magnetic resonance imaging\index{Diffusion magnetic resonance imaging} (dMRI) is a non-invasive medical imaging modality that has important uses for studying neurological disease pathology related to the anatomical network of neuronal fibers in the brain.  Advanced dMRI protocols, like diffusion spectrum imaging (DSI) and high angular resolution diffusion imaging\index{high angular resolution diffusion imaging} (HARDI) have been proven to outperform the popularly used diffusion tensor imaging by producing more accurate estimations of fiber tracts.  However, their utilization in the clinical setting is hampered by the increased number of diffusion measurements that are typically required. In order to accelerate dMRI and maintain accurate signal reconstructions, compressed sensing\index{compressed sensing} (CS) has been regularly employed in the literature.  The main ingredients of the CS framework are an appropriately chosen sampling scheme and an underlying ``sparse" representation of the data. The key idea is that, the sparser the representation, the fewer the samples needed to reconstruct the full signal with high accuracy.  

CS has been classically applied to MRI \cite{Lustig:MRM07} by subsampling in the native $k$-space (k-CS) while applying sparsifying transforms in the spatial image domain like wavelets and total-variation (TV).  For dMRI, diffusion signals are measured along different angular gradient directions in $q$-space for every point in $k$-space. Thus, to reduce the number of diffusion measurements, many methods \cite{Ning:MIA15} have exploited sparse representations in the angular domain by applying CS in $q$-space ($q$-CS).  To further accelerate dMRI, more recent methods \cite{Cheng:IPMI15,Sun:IPMI15,Mani:MRM15,Ning:Neuroimage16} combine aspects of $k$-CS and $q$-CS by subsampling jointly in ($k,q$)-space ($(k,q)$-CS). However, these methods impose sparsity on the spatial and angular domains \textit{separately}, which can lead to a less efficient representation of dMRI data and may limit the reduction of signal measurements that can be achieved in ($k,q$)-CS.


In this paper, we present a new ($k,q$)-CS\index{($k$,$q$)-CS} framework that subsamples jointly in ($k,q$)-space while analogously imposing sparsity in the \textit{joint} spatial-angular domain. Building upon the recent findings of \cite{Schwab:MICCAI16,Schwab:MIA-ArXiv17} which show increased levels of dMRI sparsity using joint spatial-angular sparse coding\index{joint spatial-angular sparse coding}, our proposed ($k,q$)-CS has the potential to further accelerate dMRI than prior methods by exploiting this underlying sparse representation.
Our main objective in this paper is to evaluate the advantages of imposing sparsity in the joint spatial-angular domain versus previous formulations that involve separate spatial and angular sparsity terms. For this reason, our focus will not yet be the optimization of sparsifying dictionaries or sensing schemes to push the limits of subsampling but first to compare the gains of our proposed model with respect to the state-of-the-art formulations for a fixed setting.

\section{Background and Prior Work}
\label{sec:prior}
\myparagraph{Compressed Sensing (CS)}
\label{sec:priorCS}
CS is a popular framework that allows for full signal recovery via irregular sampling by exploiting a sparse representation of the data \cite{CandesE2006-ICM}. In the general setting, a full signal $s$ is reconstructed from undersampled and noisy measurements $\hat{s}$ obtained through an undersampling (or sensing) matrix $\mathcal{U}$ by solving an $\ell^1$ minimization program of the form:
\begin{equation}
\label{eq:CS}
\min_{s,c} \frac{1}{2} ||\mathcal{U} s - \hat{s}||_2^2 + \lambda||c||_1,
\end{equation}
subject to the constraint that \textbf{either} $s=\Phi c$ with $\Phi$ being a sparsifying dictionary and $c$ the coefficients (\textit{synthesis}) \textbf{or} $c=\Phi^\top s$ where $\Phi^\top$ is an analysis operator applied to the signal (\textit{analysis}). Both formulations involve a sparsity prior $\|c\|_1$ in the transform domain of the signal space that is controlled by the balance parameter $\lambda\!\geq\!0$. Note however that in the typical scenario in which $\Phi$ is an overcomplete dictionary, \textit{synthesis} and \textit{analysis} CS are not equivalent models (cf. \cite{Elad2007} for a thorough discussion). In the \textit{synthesis} case, the optimization is done on the coefficient vector $c$ from which the $s$ can be synthesized while in the \textit{analysis} case $s$ is found directly. Conditions for guaranteeing recovery, which are based on the sparsity level of the signal and the mutual incoherence between $\mathcal{U}$ and $\Phi$, state that the sparser the representation and the more incoherent the sampling, the fewer measurements are required to fully reconstruct the signal.

\myparagraph{$k$-CS for MRI} 
\label{sec:priorkCS}
One of the first applications of CS has been the acceleration of MRI acquisition \cite{Lustig:MRM07}. Measurements are made in the frequency domain (called $k$-space) and the reconstruction is done in the image domain. If we denote by $\hat{s}_k$ the subsampled measurements in $k$-space and by $s_x$ the fully reconstructed image, the CS problem \eqref{eq:CS} for MRI becomes:
\begin{equation}
\label{eq:kCS}
\min_{s_x,b} \frac{1}{2} ||\mathcal{U}_k\mathcal{F} s_x - \hat{s}_k||_2^2 + \lambda||b||_1,
\end{equation}
subject to the constraint that \textbf{either} $s_x = \Psi b$ (\textit{synthesis}) \textbf{or} $b=\Psi^\top s_x$ (\textit{analysis}), where $\mathcal{F}$ is the Fourier Transform, $\mathcal{U}_k \in \mathbb{R}^{K\times V}$ is the undersampling $k$-space matrix, $K$ is the number of samples and $V$ is the total number of voxels with $K\leq V$. Here $\Psi \in \mathbb{R}^{V \times N_\Psi}$ is typically a dictionary of $N_\Psi$ atoms defined on the image domain (\eg Wavelets or Hadamard) and $\Psi^\top$ is a sparsifying transform either associated to those dictionaries or to other operators such as the gradient in the case of total variation (TV) regularization. This last choice in the \textit{analysis} formulation ($b = \Psi^\top s_x$) is a common model for sparse MRI reconstruction \cite{Lustig:MRM07}:
\begin{equation}
\label{eq:kCSanalysis}
\min_{s_x} \frac{1}{2} ||\mathcal{U}_k\mathcal{F} s_x - \hat{s}_k||_2^2 + \lambda||\Psi^\top s_x||_1.
\end{equation}

\myparagraph{$q$-CS for dMRI}
\label{sec:priorqCS}
The structure of dMRI is significantly more complex than that of traditional MRI, whereby for each $k$-space measurement, a set of $G$ (angular) diffusion measurements are acquired in the analogous $q$-space.
Diffusion signals are traditionally viewed voxel-wise in the image domain (after $k$-space reconstruction) as a matrix $S_{x,q}=[s_1,\dots, s_V]^\top \in \mathbb{R}^{V\times G}$, where $s_v \in \mathbb{R}^G$ is the diffusion signal in voxel $v$. $q$-CS has been used extensively in the literature \cite{Ning:MIA15}, each new treatment testing a new sparsifying angular dictionary or sampling scheme. Traditionally formulated as in \eqref{eq:CS} for each voxel $v$,
$q$-CS is more frequently solved for all voxels simultaneously as:
\begin{equation}
\label{eq:qCS}
\min_{S_{x,q},A} \frac{1}{2} || S_{x,q} \mathcal{U}_q^\top - \hat{S}_{x,q}||_F^2 + \lambda ||A||_1,
\end{equation}
subject to the constraint that \textbf{either} $S_{x,q} =A \Gamma^T$ (\textit{synthesis}) \textbf{or} $A = S_{x,q}\Gamma$ (\textit{analysis}), where $\hat{S}_{x,q} = [\hat{s}_1,\dots, \hat{s}_V]^\top \in \mathbb{R}^{V\times Q}$ are the measured $q$-space signals $\hat{s}_v \in \mathbb{R}^Q$ at each voxel $v$, $\mathcal{U}_q \in \mathbb{R}^{Q \times G}$ is an undersampling matrix in $q$-space with $Q \leq G$, and $A = [a_1, \dots, a_V]^\top \in \mathbb{R}^{V \times N_\Gamma}$ is the matrix of angular coefficients for an angular $q$-space dictionary $\Gamma \in\mathbb{R}^{G\times N_\Gamma}$ with $N_\Gamma$ atoms.  Prior work \cite{Ning:MIA15} has explored the construction of many sparsifying dictionaries $\Gamma$ related to estimating orientation distribution functions and so the constraint $S_{x,q} = A\Gamma^\top$ is most commonly used, resulting in the \textit{synthesis} formulation:
\begin{equation}
\label{eq:qCSsynthesis}
\min_{A} \frac{1}{2} || A\Gamma^\top \mathcal{U}_q^\top - \hat{S}_{x,q}||_F^2 + \lambda ||A||_1.
\end{equation}

\begin{figure}
\centering
\includegraphics[width=1\linewidth,trim=50 0 30 0, clip]{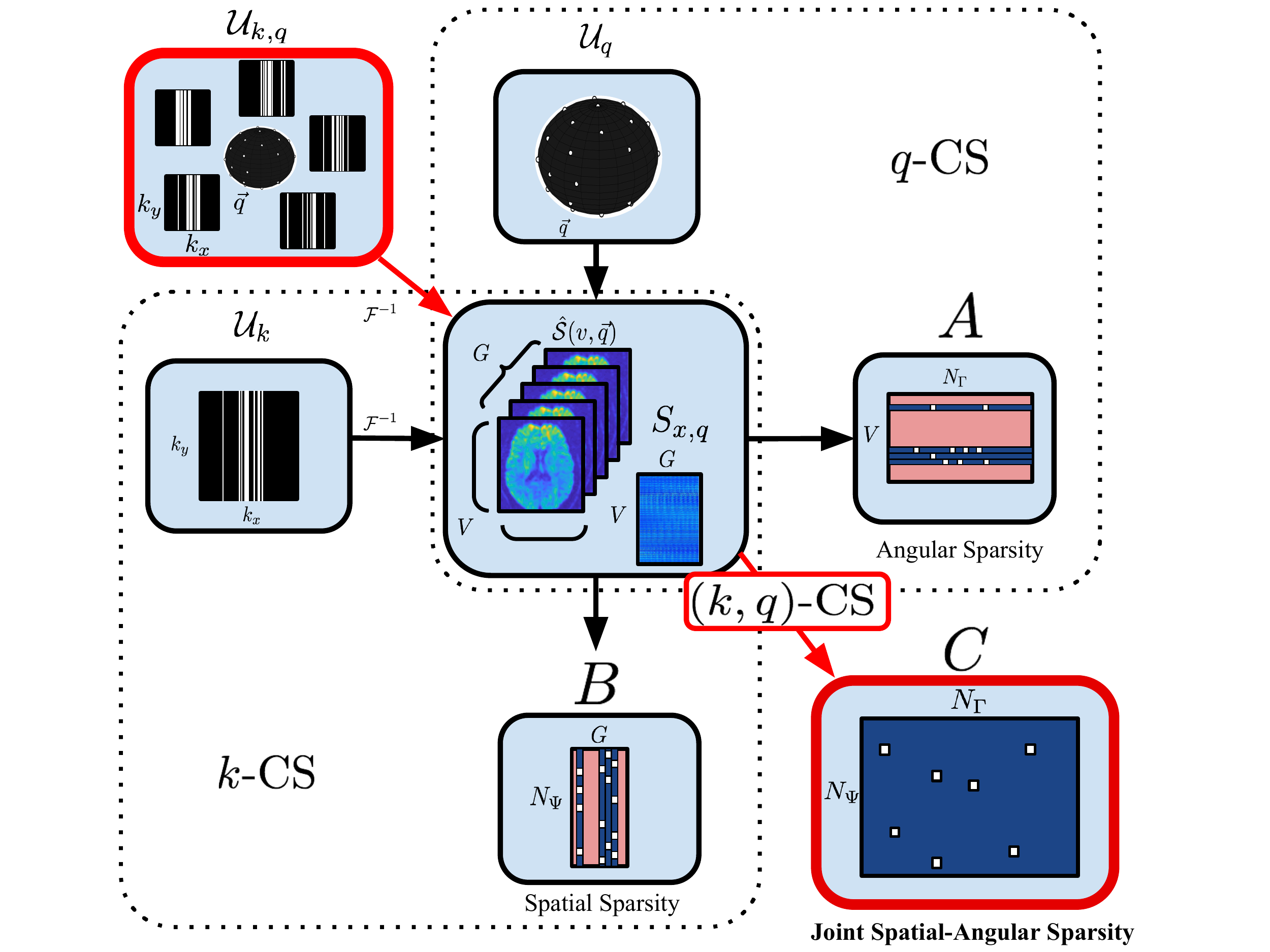}
\caption{Diagram of $k$-CS, $q$-CS, and $(k,q)$-CS with domains of sensing (top left) and sparsity (bottom right).  State-of-the-art methods subsample jointly in ($k,q$)-space with $\mathcal{U}_{k,q}$ but then \textit{add} separate spatial, $B$ (bottom), and angular, $A$ (right), sparsity combining $k$- and $q$-CS.  Instead, we propose to enforce sparsity in the joint spatial-angular domain, $C$ (bottom-right), resulting in a natural unified framework for ($k,q$)-CS that allows a reduced number of samples via increased levels of joint sparsity.}
\label{fig:diagram}
\end{figure}

\myparagraph{$(k,q)$-CS for dMRI}
\label{sec:priorkqCS} 
A logical advancement to further accelerate dMRI is to additionally subsample in $k$-space.  State-of-the-art methods like \cite{Cheng:IPMI15,Sun:IPMI15,Mani:MRM15,Ning:Neuroimage16} have been applied to many dMRI protocols testing various combinations of dictionaries and sensing schemes. Interestingly, all of them can be formulated as particular cases of the following problem, which combine $k$-CS \eqref{eq:kCS} and $q$-CS \eqref{eq:qCS}:
\begin{equation}
\label{eq:kqCSmat_sep}
\min_{S_{x,q},A,B} \frac{1}{2}||\mathcal{U}_{k,q}(\mathcal{F}S_{x,q})  - \hat{S}_{k,q}||_F^2 + \lambda_1||A||_1 + \lambda_2||B||_1. 
\end{equation}
subject to the constraints $S_{x,q} = A\Gamma^\top $ (\textit{synthesis} as in \eqref{eq:qCSsynthesis}) and $B=\Psi^\top S_{x,q}$ (\textit{analysis} as in \eqref{eq:kCSanalysis}). 
The sensing scheme $\mathcal{U}_{k,q}$ is now a joint $(k,q)$ subsampling operator (cf. Fig.~\ref{fig:diagram} and Sec.~\ref{sec:transforms} for a discussion) and $\hat{S}_{k,q} \in \mathbb{R}^{K \times Q}$ are the subsampled measurements in $(k,q)$-space.
As a critical point of distinction, in \eqref{eq:kqCSmat_sep} the sparsity prior is imposed on two separate domains: the angular dictionary coefficients $A \in \mathbb{R}^{V\times N_\Gamma}$ at each voxel and the spatial transform coefficients $B \in \mathbb{R}^{G\times N_\Psi}$ for each gradient direction. The sparsity in these domains, if measured by the $\ell_0$ ``norm", is inherently limited by the size of the dMRI data ($V$,$G$) since, for non-zero $q$-space signals at all voxels $||A||_0\geq V$, and for non-zero $k$-space images for each gradient direction $||B||_0 \geq G$, resulting in a total spatial plus angular sparsity of $||A||_0 + ||B||_0 \geq V+G$. This limitation of sparsity may eventually impact the possible reduction in sampling rate for ($k,q$)-CS as we will show empirically in our experiments in Section~\ref{sec:experiments}. 

To overcome this limitation, we propose an alternative and unified ($k,q$)-CS formulation which instead imposes sparsity on a natural \textit{joint} spatial-angular domain, which we call $C$, defined formally in the next section. Fig.~\ref{fig:diagram} depicts a full schematic summary of the domains of sampling in $k$-CS, $q$-CS, and the joint ($k,q$)-CS, and the associated sparsity priors in the spatial domain ($B$), angular domain ($A$) and the joint spatial-angular ($C$) domain. As motivated by Fig.~\ref{fig:diagram}, while $A$ is row-sparse ($||A||_0 \geq V$), and $B$ is column sparse ($||B||_0\geq G$), $C$ has no \textit{a priori} structured sparsity ($||C||_0\geq 1$), meaning that our formulation has the potential to achieve greater sparsity levels and therefore higher subsampling rates within ($k,q$)-CS than the state of the art.


\section{Methods}
\label{sec:methods}
\subsection{($k,q$)-CS for dMRI with Joint Spatial-Angular Sparsity}
\label{sec:proposed_framework}
We propose a new ($k,q$)-CS model for dMRI involving a single joint spatial-angular sparsity prior instead of separate spatial and angular sparsity terms as in \eqref{eq:kqCSmat_sep}.  This idea stems from the spatial-angular sparse coding approach for dMRI proposed in \cite{Schwab:MICCAI16,Schwab:MIA-ArXiv17}, which was shown to result in much sparser representations than separate spatial and angular dictionaries. %
%
Specifically, instead of using a voxel-wise viewpoint of a dMRI signal written in an angular dictionary for every voxel, we consider the full global signal $s_{x,q} \in \mathbb{R}^{VG}$, the stacking of each $s_x$ for every $q$-space point, and a measured subsampled signal in $(k,q)$-space $\hat{s}_{k,q}\in \mathbb{R}^{KQ}$, such that $\hat{s}_{k,q} = \mathcal{U}_{k,q}(\mathcal{F} s_{x,q})$ where the Fourier transform $\mathcal{F}$ is applied to each spatial component and $\mathcal{U}_{k,q} \in \mathbb{R}^{KQ \times VG}$ is the $(k,q)$ sensing matrix. Then we can write the global ($k,q$)-CS in vector form, analogous to the general setting in \eqref{eq:CS}:
\begin{equation}
\label{eq:kqCSvec}
\min_{s_{x,q},c} \frac{1}{2} ||\mathcal{U}_{k,q}(\mathcal{F} s_{x,q}) - \hat{s}_{k,q}||_2^2 + \lambda||c||_1,
\end{equation}
subject to the constraint that \textbf{either} $s_{x,q} = \Phi c$ (\textit{synthesis}) \textbf{or} $c=\Phi^\top s_{x,q}$ (\textit{analysis}). Notice that \eqref{eq:kqCSvec} has a direct statistical interpretation as a reconstruction under a sparsity prior with respect to the dictionary $\Phi \in \mathbb{R}^{VG \times N_\Phi}$. However, numerically solving such an optimization problem is largely intractable due to the size of dMRI data ($|s_{x,q}| = VG \approx 100^4$) and the resulting huge size of $\Phi$. 

To overcome this difficulty, we propose to impose additional structure on $\Phi$. Following \cite{Schwab:MICCAI16,Schwab:MIA-ArXiv17} one can choose $\Phi$ to be separable over the spatial and angular domains resulting in the Kronecker\index{Kronecker} dictionary $\Phi = \Gamma \otimes \Psi$, where $\Psi$ and $\Gamma$ are, as above, spatial and angular dictionaries, respectively. Then, we can rewrite \eqref{eq:kqCSvec} in an equivalent matrix form as:
\begin{equation}
\label{eq:kqCSmat}
\min_{S_{x,q},C} \frac{1}{2}||\mathcal{U}_{k,q}( \mathcal{F}S_{x,q}) - \hat{S}_{k,q}||_F^2 + \lambda||C||_1,
\end{equation}
subject to the constraint that \textbf{either} $S_{x,q} = \Psi C \Gamma^\top$ (\textit{synthesis}) \textbf{or} $C =  \Psi^\top S_{x,q} \Gamma$ (\textit{analysis}). In fact, substituting also the constraints from $k$-CS \eqref{eq:kCS} and $q$-CS \eqref{eq:qCS}, a separable spatial-angular dictionary allows us to have two additional constraint options: (1) $S_{x,q} = A\Gamma^\top$ \textbf{and} $C = \Psi^\top A$ (\textit{analysis}-\textit{synthesis}) \textbf{or} (2) $S_{x,q} = \Psi B$ \textbf{and} $C = B\Gamma$ (\textit{synthesis}-\textit{analysis}).

Notice that, in contrast to the state-of-the-art formulation in \eqref{eq:kqCSmat_sep}, our formulation only involves one penalty term that imposes sparsity on the \textit{joint} spatial-angular coefficient domain $C \in \mathbb{R}^{N_\Psi \times N_\Gamma}$ of the global dictionary $\Gamma \otimes \Psi$ (cf. Fig. \ref{fig:diagram}). The sparsity of this domain is \textit{a priori} not limited by the size of the data and so this joint model can lead to sparser representations of typical dMRI signals than summing separate spatial and angular terms.  In the next section, we present an algorithm to efficiently solve the proposed ($k,q$)-CS formulation. 

\subsection{Efficient Algorithm to Solve ($k,q$)-CS}
\label{sec:KronSFISTA}

Prior work such as \cite{Cheng:IPMI15,Ning:Neuroimage16,Sun:IPMI15} each solve \eqref{eq:kqCSmat_sep} using the Split-Bregman/Alternating Direction Method of Multipliers (ADMM) algorithm and divide the reconstruction per voxel. Alternatively, we propose an efficient algorithm to solve ($k,q$)-CS globally for large-scale dMRI data which can easily be applied to both the prior formulation \eqref{eq:kqCSmat_sep} and our proposed formulation \eqref{eq:kqCSmat}.

We begin by taking care of the constraints to eliminate variables and simplify the problems. For \eqref{eq:kqCSmat_sep}, we substitute the prior methods' selected constraints $S_{x,q} = A\Gamma^\top$ and $B = \Psi^\top S_{x,q} = \Psi^\top A \Gamma^\top$ to get:
\begin{equation}
\label{eq:kqCSmat_sep_A}
\min_{A} \frac{1}{2}||\mathcal{U}_{k,q}(\mathcal{F}A \Gamma^\top)  - \hat{S}_{k,q}||_F^2 + \lambda_1||A||_1 + \lambda_2||\Psi^\top A \Gamma^\top||_1. 
\tag{Prior}
\end{equation}
In order to directly compare our proposed framework \eqref{eq:kqCSmat} with \eqref{eq:kqCSmat_sep_A} in terms of variable $A$, we substitute $S_{x,q} = A\Gamma^\top$ and $C=\Psi^\top A$ (\textit{analysis}-\textit{synthesis}) to get:
\begin{equation}
\label{eq:kqCSmat_A}
\min_A  \frac{1}{2}||\mathcal{U}_{k,q}(\mathcal{F} A \Gamma^\top)  - \hat{S}_{k,q}||_F^2 + \lambda|| \Psi^\top A||_1.
\tag{SAAS}
\end{equation}
We call this formulation \textbf{Spatial-Angular} \textbf{Analysis-Synthesis} \textbf{(SAAS)} due to the resulting \textit{analysis} formulation for the spatial domain and \textit{synthesis} formulation for the angular domain. While these substitutions mask the domains of sparsity by using a common variable $A\in \mathbb{R}^{V\times N_\Gamma}$, note that the proposed formulation \eqref{eq:kqCSmat_A} still imposes sparsity on the joint spatial-angular domain in contrast to the separate spatial and angular sparsity terms of \eqref{eq:kqCSmat_sep_A}. 

The Fast Iterative Shrinkage-Thresholding Algorithm (FISTA) \cite{Beck2009} has been well studied for solving $\ell^1$ \textit{synthesis} minimization problems such as \eqref{eq:CS}, where the proximal operator of $||c||_1$ is the well-known shrinkage function.
However, in the \textit{analysis} setting, the proximal operator of a linearly transformed variable such as $||\Psi^\top A||_1$ in \eqref{eq:kqCSmat_A} and $||\Psi^\top A \Gamma^\top||_1$ in \eqref{eq:kqCSmat_sep_A} is not directly computable. There are multiple ways to overcome this. In particular, \cite{Tan:TSP14} proposes a method that applies FISTA to a relaxed \textit{smooth} problem, coined Smooth FISTA (SFISTA). In what follows, we adapt SFISTA to the separable Kronecker matrix setting in order to solve \eqref{eq:kqCSmat_A} and \eqref{eq:kqCSmat_sep_A}. 

First, \eqref{eq:kqCSmat_A} is reformulated by introducing the auxiliary linear constraint $Z=\Psi^\top A$ and the unconstrained relaxed optimization becomes:
\begin{equation}
\label{eq:l1_separable_relaxedanalysis}
\min_{A,Z} \frac{1}{2} ||\mathcal{U}_{k,q}(\mathcal{F}A \Gamma^\top )  - \hat{S}_{k,q}||_F^2 + \lambda||Z||_1 + \frac{\rho}{2}||Z - \Psi^\top A||_F^2.
\end{equation}
Let $f(A) \equiv ||\mathcal{U}_{k,q}(\mathcal{F}A \Gamma^\top )  - \hat{S}_{k,q}||_F^2$. Since $f$ does not depend on $Z$, we can pass the minimization with respect to $Z$ to the last two terms. Define
$g_\mu(X) \equiv \min_Z ||Z||_1 + \frac{1}{2\mu}||Z - X ||_F^2$. Then \eqref{eq:l1_separable_relaxedanalysis} is equivalent to
\begin{equation}
\label{eq:fplusg}
\min_{A}f(A) +\lambda g_\frac{\lambda}{\rho}(\Psi^\top A).
\end{equation}
Here 
$g_\mu$ is the Moreau envelope of the $\ell^1$ norm which can be shown to equal the Huber function given by $\mathcal{H}_\mu(x)$ = $\frac{1}{2\mu^2} x^2$ if $|x| < \mu$ and $|x| - \frac{\mu}{2}$ otherwise. We can now apply FISTA to the smooth \eqref{eq:fplusg} by taking an accelerated gradient descent using
\begin{align}
&\nabla f(A) = \mathcal{F}^{-1}\mathcal{U}_{k,q}^* ( \mathcal{U}_{k,q}(\mathcal{F} A \Gamma^\top))\Gamma - \mathcal{F}^{-1}\mathcal{U}_{k,q}^* (\hat{S}_{k,q}) \Gamma \\
&\nabla g_{\frac{\lambda}{\rho}}(\Psi^\top A) = \frac{\rho}{\lambda} \Psi (\Psi^\top A - \textnormal{shrink}_{\frac{\lambda}{\rho}} (\Psi^\top A ))
\end{align}
where $\mathcal{U}^*_{k,q}$ is the operator that restores the subsampled signal to full size by zeroing out unsampled indices of the full signal (cf. Sec.~\ref{sec:transforms} for a discussion).  The proposed Kronecker SFISTA (Kron-SFISTA) is presented in Algorithm~\ref{alg:KronSFISTA}.
\vspace{-4ex}
\begin{algorithm}
\caption{Kron-SFISTA for SAAS model ($k,q$)-CS}
\begin{algorithmic}
\label{alg:KronSFISTA}
\STATE Choose: $\lambda,\rho,\epsilon$.
\STATE Initialize: $i=1,Y_1 = S_0 = \mathbf{0}$, $n_1 = 1, L\geq \lambda_{\textnormal{max}}(\Gamma^\top \Gamma) + \rho \lambda_{\textnormal{max}}(\Psi\Psi^\top)$.
\WHILE {error $> \epsilon$}
\STATE 1: $A_i = Y_i - (\nabla f(Y_i) + \lambda\nabla g_{\lambda/\rho}(\Psi^\top Y_i)) /L;$
\STATE 2: $n_{i+1} = \frac{1}{2}(1 + \sqrt{1 + 4n_i^2});$
\STATE 3: $Y_{i+1} = A_{i} + \frac{n_i-1}{n_{i+1}}(A_{i} - A_{i-1});$
\STATE 4: $i \leftarrow i+1;$
\ENDWHILE
\STATE Return: $\hat{A}$. 
\STATE Reconstruct: $\hat{S}_{x,q} = \hat{A}\Gamma^\top$.
\end{algorithmic}
\end{algorithm}
\vspace{-4ex}

According to \cite{Tan:TSP14}, we choose stepsize $L\geq \lambda_{\textnormal{max}}(\Gamma^\top \Gamma) + \rho \lambda_{\textnormal{max}}(\Psi\Psi^\top)$ to guarantee convergence, where $\lambda_{\textnormal{max}}(X)$ is the max eigenvalue of $X$. The parameter $\rho$ is gradually increased using parameter continuation \cite{Tan:TSP14} for ensuring convergence. The trade-off parameter $\lambda$, dictates the level of sparsity of $\Psi^\top A$. A large value of $\lambda$ will result in a very sparse representation at the expense of reconstruction accuracy, while a small value of $\lambda$ may result in over-fitting the sampled data at the expense of reconstruction accuracy of unseen data. Therefore, in our experiments we vary the level of $\lambda$ and select the value that leads to a minimal reconstruction error. The efficiency of Kron-SFISTA over the traditional SFISTA can be viewed in the same vein as for Kron-FISTA analyzed in \cite{Schwab:MIA-ArXiv17}.

As an alternative to the frequently used Split-Bregman, Kron-SFISTA can also be easily applied to \eqref{eq:kqCSmat_sep_A} by solving:
\begin{equation}
\min_{A} f(A) + \lambda_1||A||_1 + \lambda_2 g_{\frac{\lambda_2}{\rho_2}} (\Psi^\top A\Gamma^\top).
\end{equation}
Step 1 in Alg.~\ref{alg:KronSFISTA} becomes $A_i = \textnormal{shrink}_{\lambda_1}( Y_i  - (\nabla f(Y_i) + \lambda_2 \nabla g_{\frac{\lambda_2}{\rho_2}} (\Psi^\top Y_i\Gamma^\top))/L)$ with $\nabla g_{\frac{\lambda_2}{\rho_2}}(\Psi^\top A\Gamma^\top) = \frac{\rho_2}{\lambda_2} \Psi (\Psi^\top A\Gamma^\top - \textnormal{shrink}_{\frac{\lambda_2}{\rho_2}} (\Psi^\top A\Gamma^\top ))\Gamma$.
This provides an efficient global algorithm to solve $(k,q)$-CS for any large-scale dMRI data. 
\section{Experiments}
\label{sec:experiments}
 The main objective of the current work is to directly compare the reconstruction accuracy of \eqref{eq:kqCSmat_A} and \eqref{eq:kqCSmat_sep_A} for various rates of subsampling.  We postpone optimizing the amount of subsampling to future work and therefore explore somewhat classical choices of spatial and angular dictionaries/transforms and sensing schemes previously tested in the literature in Sec.~\ref{sec:transforms} with experimental results on phantom and real HARDI brain data in Sec.~\ref{sec:phantom} and Sec.~\ref{sec:real}.
 
\subsection{Spatial-Angular Transforms and ($k,q$) Subsampling Schemes}
\label{sec:transforms}

\myparagraph{Spatial Transform $\Psi^\top$} 
For spatial transform $\Psi^\top$, we consider in our experiments two popularly used transforms for $k$-CS: Haar wavelets and the finite difference (gradient) operator $\nabla = [\partial_x, \partial_y, \partial_z]$. 
In the case of the gradient transform, we consider the norm given by $||\nabla(X)||_{2,1} = ||\sqrt{|\partial_x X|^2 + |\partial_y X|^2 + |\partial_z X|^2}||_1$, known as isotropic TV (isoTV)\footnote{SFISTA must be changed slightly to incorporate the $||\cdot||_{2,1}$ proximal operator shrink$^{2,1}_\mu (X) = \frac{X}{||X||_{2,\cdot}}\max(||X||_{2,\cdot} - \mu,0)$ \cite{Goldstein:SIAM09}, where $||X||_{2,\cdot}$ indicates taking the 2-norm of the columns of $X$.  Its Moreau envelope is $g^{2,1}_\mu(X) \equiv \min_Z ||Z||_{2,1} + \frac{1}{2\mu}||Z - X||_F^2 = \frac{1}{2\mu^2} ||X||_{2,\cdot}^2$ if $||X||_{2,\cdot} < \mu$ and $||X||_{2,\cdot} - \frac{\mu}{2}$ otherwise.}. These transforms have been classically used to sparsely represent MRI images.

\myparagraph{Angular Dictionary $\Gamma$}
The choice of angular dictionary $\Gamma$ depends on the $q$-space acquisition protocol of the data. For example, $\Gamma$ must be chosen to model Cartesian sampled $q$-space signals for DSI, and multi-shell $q$-space signals with a radial component for multi-shell HARDI.  It is important to note our framework is general to any $q$-space acquisition protocol with appropriate choice of $\Gamma$. In our experiments we use single-shell HARDI data and choose the over-complete spherical ridgelet (SR) dictionary \cite{TristanVega:MICCAI11}, which has been shown to sparsely model HARDI signals. With this comes the spherical wavelet (SW) dictionary for which we can estimate orientation distribution functions (ODFs) from the SR coefficients. With our choice of parameters, this results in $N_\Gamma = 1169$ atoms from which we may choose any subset greater than $G$ for an overcomplete dictionary.

\myparagraph{Joint $(k,q)$ Subsampling Scheme $\mathcal{U}_{k,q}$}
We experiment with different subsampling schemes in the joint $(k,q)$ domain. Along the lines of our separable sparsifying dictionaries, we first consider separable sensing schemes for which $\mathcal{U}_{k,q} = \mathcal{U}_{k} \otimes \mathcal{U}_{q}$ and $\mathcal{U}_{k,q}(\mathcal{F}A \Gamma^\top ) = \mathcal{U}_{k} \mathcal{F}A \Gamma^\top \mathcal{U}_{q}^\top $ where $\mathcal{U}_k \in \mathbb{R}^{K \times V}$ and $\mathcal{U}_q \in \mathbb{R}^{Q \times G}$ can be taken to be matrices. Algorithmically, this is straight forward to compute and $\mathcal{U}^*_{k,q}$ is simply $\mathcal{U}^\top_{k,q} = \mathcal{U}_k^\top \otimes \mathcal{U}_q^\top$. In our first set of experiments on phantom data (Sec.~\ref{sec:phantom}), we take $\mathcal{U}_k$ to be a random subsampling matrix with a well-known variable radial density favoring low frequencies and $\mathcal{U}_q$ a quasi-uniform random subsampling on the sphere. However, separable sensing strategies may not fully exploit the potentialities of $(k,q)$ subsampling as some experiments in \cite{Mani:MRM15} show. 

In the second set of experiments on real data (Sec.~\ref{sec:real}), we implement non-separable sensing schemes $\mathcal{U}_{k,q}$ in which a different $k$-space sampling is used for each sampled $q$-space point. In this case, $\mathcal{U}^*_{k,q}$ is the operator that restores the subsampled signal to full size by zeroing out unsampled indices of the full signal.  Our implementation of Kron-SFISTA has the benefit of being able to easily handle this non-separable sensing operator, but this is not straight forward in alternative algorithmic formulations such as a Kron-ADMM \cite{Schwab:MIA-ArXiv17}, for example. For $k$-space sensing, to comply with the constraints of a physical scanner, we choose a commonly used $k$-space sampling scheme of constant lines along the $k_y$ direction. The $k_x$ location of the line samples were chosen randomly with respect to a variable density function centered around the zero-frequency location. We generated a different random $k$-space sampling for each chosen $q$-space point. See Fig.~\ref{fig:diagram} (top left) for a visualization of our joint ($k,q$) sampling $\mathcal{U}_{k,q}$.

Intuitively, non-separable sensing increases the range of uniquely sampled points and the level of randomness  which are beneficial in CS. In preliminary experiments comparing separable to non-separble sensing for real HARDI data, we saw variable improvement of reconstruction accuracy for both \eqref{eq:kqCSmat_sep_A} and \eqref{eq:kqCSmat_A}, depending on the choices of $\mathcal{U}_k$ and $\mathcal{U}_q$, which motivates our use of non-separable sensing in Sec.~\ref{sec:real}. However, the theoretical analysis of coherence and sparsity with respect to sensing and dictionaries in the CS paradigm is beyond the scope of the current work.
\begin{figure}
\centering
\includegraphics[width=1\linewidth,height=.5\linewidth,trim=250 20 250 20,clip]{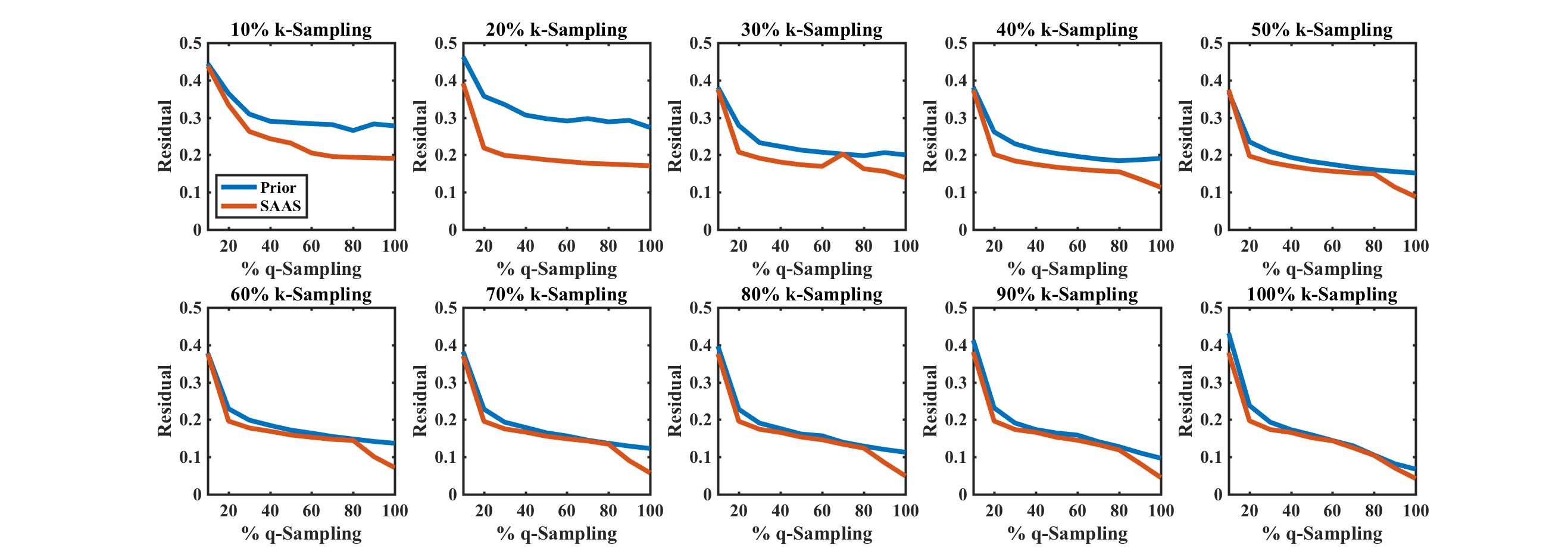}
\caption{Residual error vs. percentage of ($k,q$) subsampling of the 2D Phantom HARDI data using isoTV and SR for \eqref{eq:kqCSmat_A} (red) and \eqref{eq:kqCSmat_sep_A} (blue). \eqref{eq:kqCSmat_A} provides more accurate reconstruction, especially at lower levels of ($k,q$) subsampling (top left plots).}
\label{fig:kqsub}
\end{figure}
\begin{figure}
\centering
\includegraphics[width=.32\linewidth]{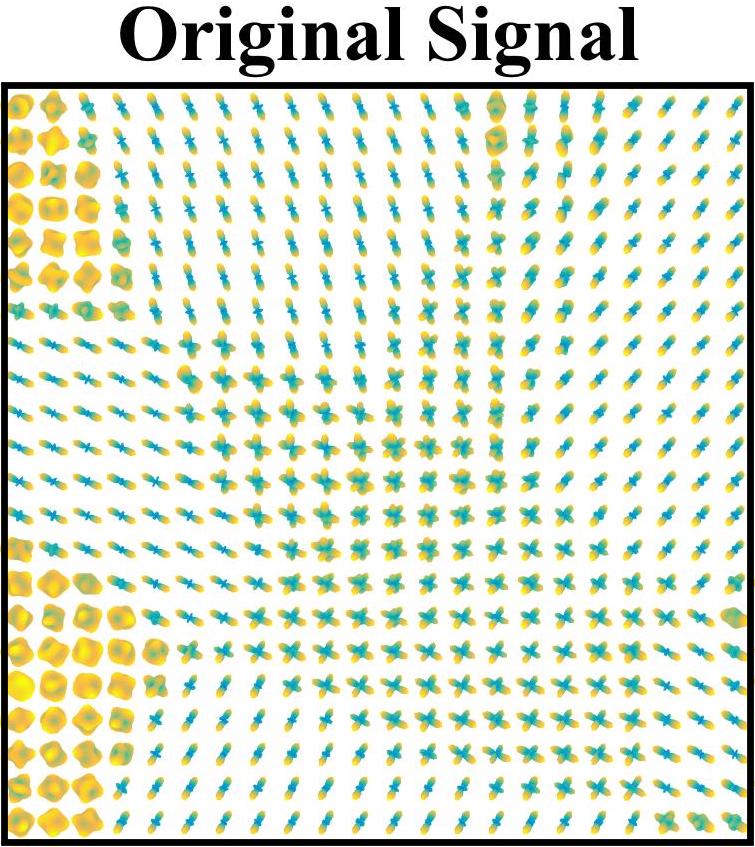}
\includegraphics[width=.32\linewidth]{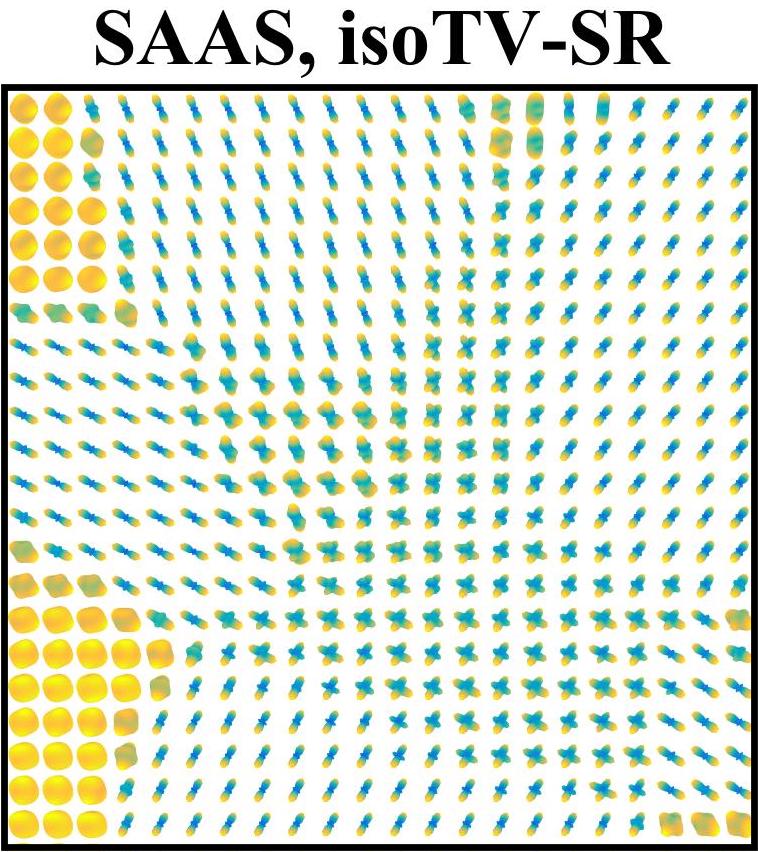}
\includegraphics[width=.32\linewidth]{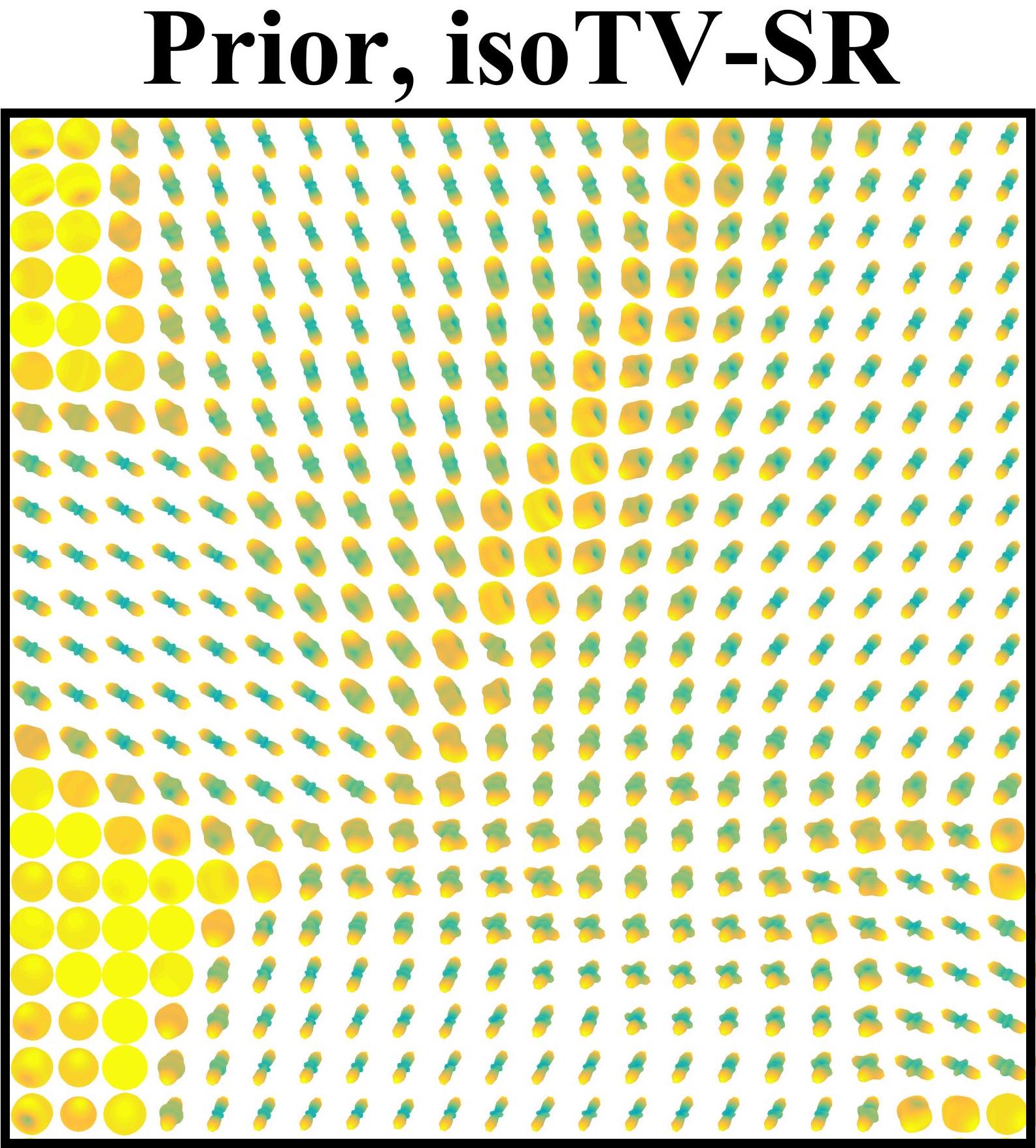}\\
\includegraphics[width=.32\linewidth]{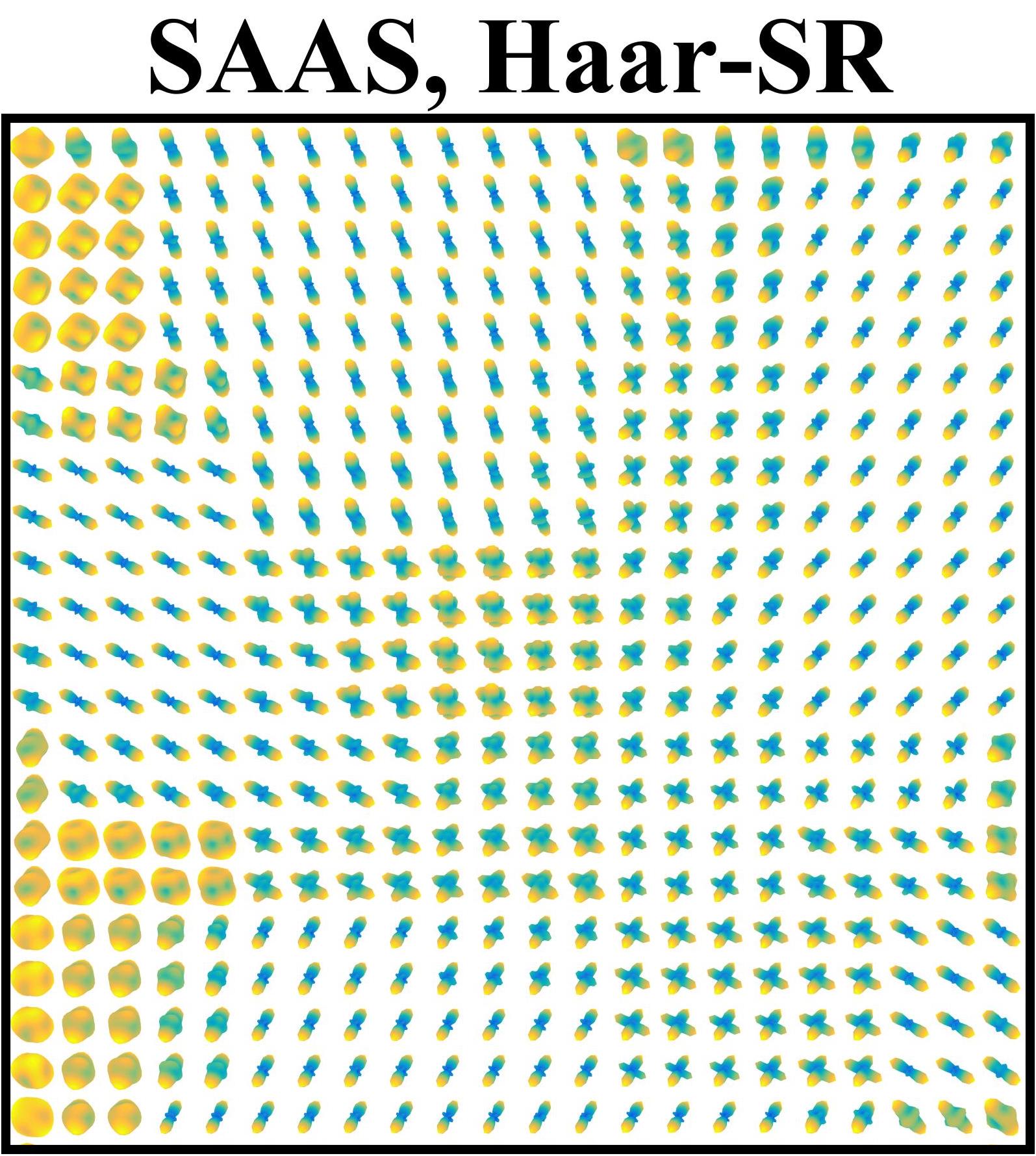}
\includegraphics[width=.32\linewidth]{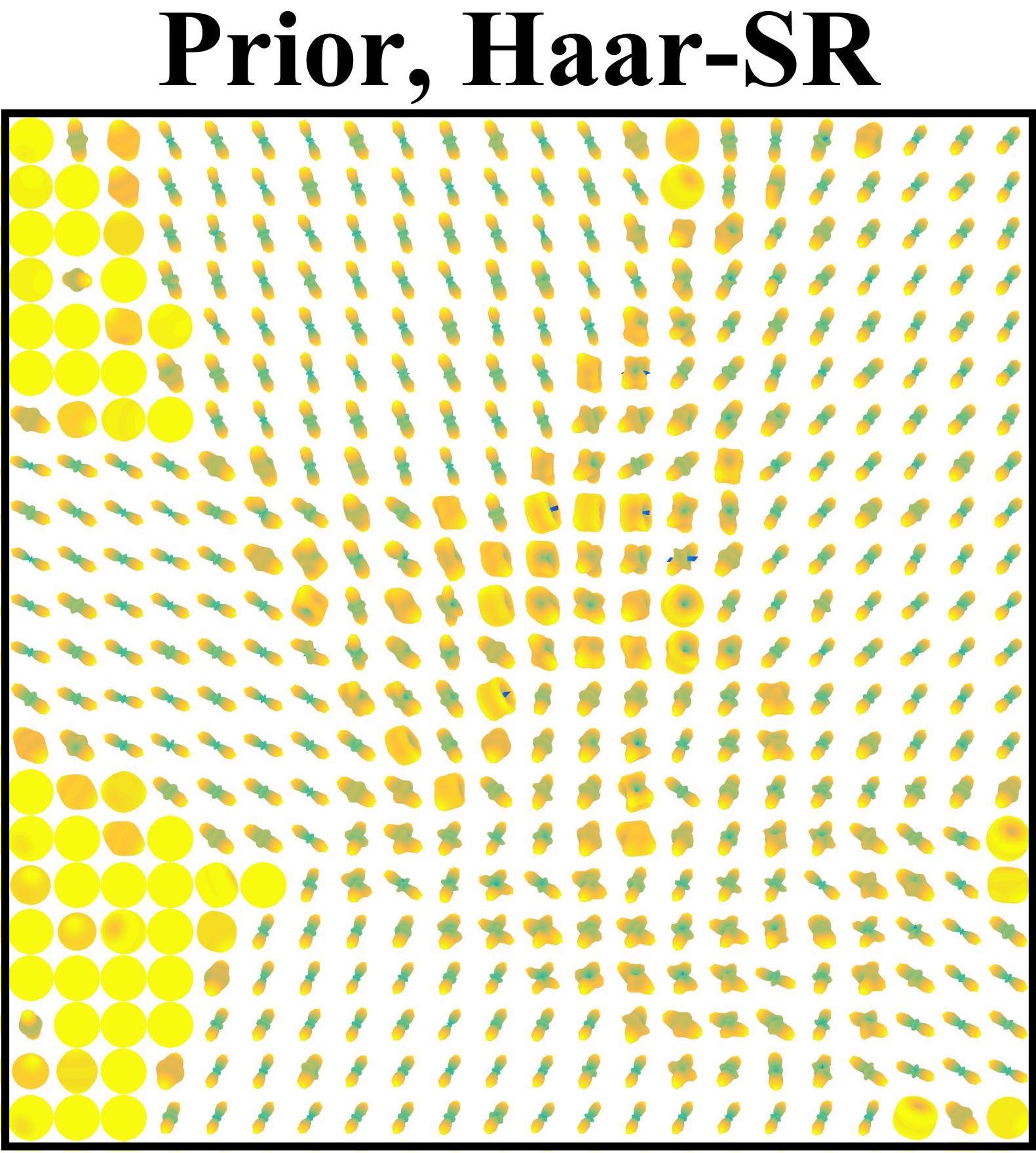}
\caption{Estimation of ODFs from reconstructed phantom signals compared to the original fully sampled signal using the proposed \eqref{eq:kqCSmat_A} and \eqref{eq:kqCSmat_sep_A}.  Each is reconstructed from $4\%$ total ($k,q$) measurements, keeping $20\%$ $k$-space samples and $20\%$ $q$-space samples.  It is apparent that the prior model is unable to accurately reconstruct crossing fiber signal in the middle of the image.  It is also evident that isoTV outperforms Haar.}
\label{fig:kqsubqual}
\end{figure}
\subsection{($k,q$)-CS Results for Phantom HARDI Data}
\label{sec:phantom}
First, we applied our methods on the ISBI 2013 HARDI Reconstruction Challenge Phantom dataset\footnote{\url{http://www.hardi.epfl.ch/static/events/2013\_ISBI/testing\_data.html}}, a $V\!=\!50\!\times\!50\!\times\!50$ volume with $G\!=\!64$ gradient directions ($b = 3000$ s/mm$^2$) and SNR $=30$, consisting of 20 phantom fibers crossing within an inscribed sphere. We experimented on a middle 2D $50\!\times\!50$ slice of this data. In this experiment, we vary the percentage of subsampling in both $k$- and $q$-space, ranging from $10\%$ to $100\%$ of the original phantom HARDI signal in each domain, resulting in a combined total of $1\%$ to $100\%$ of the full signal. Then, we compare our reconstructed signal $\hat{S}_{x,q}$ with the original full signal, $S_{x,q}$, by calculating residual error $||\hat{S}_{x,q} - S_{x,q}||^2_2 /|| S_{x,q}||^2_2$.  As a note, this phantom data has been pre-transformed to the spatial domain and so to test $k$-space subsampling, we retrospectively transform the data back to $k$-space using the Fourier transform prior to experiments. 

In Fig.~\ref{fig:kqsub} are the quantitative reconstruction results of our proposed \eqref{eq:kqCSmat_A} ($k,q$)-CS compared to \eqref{eq:kqCSmat_sep_A}. Each subplot presents a fixed $k$-space subsampling percentage, while the percentage of $q$ subsampling varies along the x-axis. Kron-SFISTA took $\sim$15-30 min to complete for a single run over multiple $\lambda$. We can see improvements of reconstruction accuracy for our proposed method especially in the desired low range of $20\%$ $k$ subsampling and $20\%$ $q$ subsampling, \ie\ $500$ frequency measurements and $12$ gradient directions, keeping a total of $4\%$ of samples (see second plot in first row of Fig.~\ref{fig:kqsub}).

We show the ODFs estimated from the reconstructed phantom signal for this $4\%$ sampling rate in Fig.~\ref{fig:kqsubqual} comparing the results of using isoTV versus Haar wavelets. 
We notice that \eqref{eq:kqCSmat_sep_A} is unable to reconstruct the complex crossing fiber ODFs in the middle region of the image at this low level of sampling. Alternatively \eqref{eq:kqCSmat_A} provides more accurate reconstructions of the entire dataset with isoTV well outperforming Haar wavelets.  

\subsection{($k,q$)-CS Results for Real HARDI Brain Data}
\label{sec:real}
\begin{figure}[H]
\centering
\includegraphics[width=.49\linewidth]{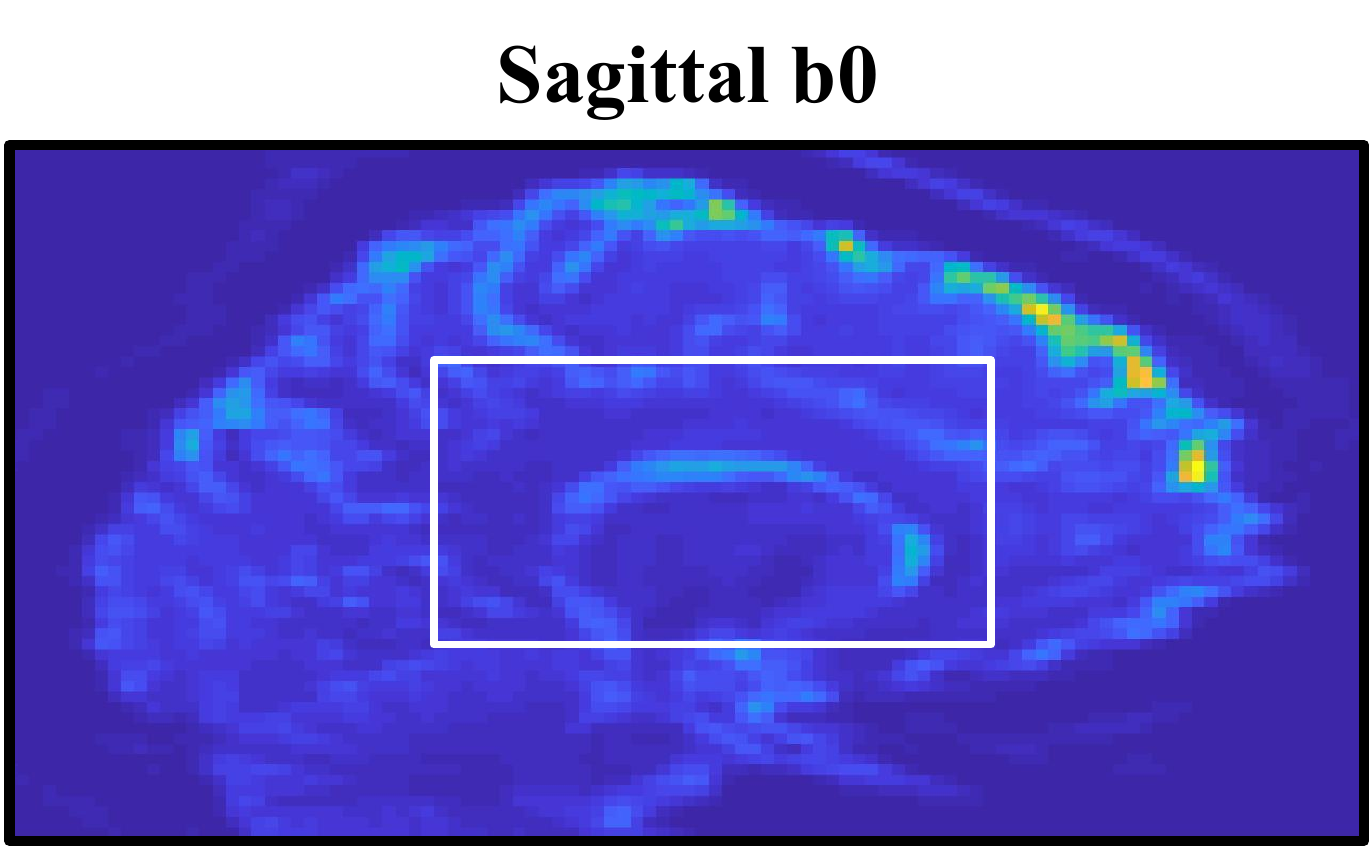}
\includegraphics[width=.49\linewidth]{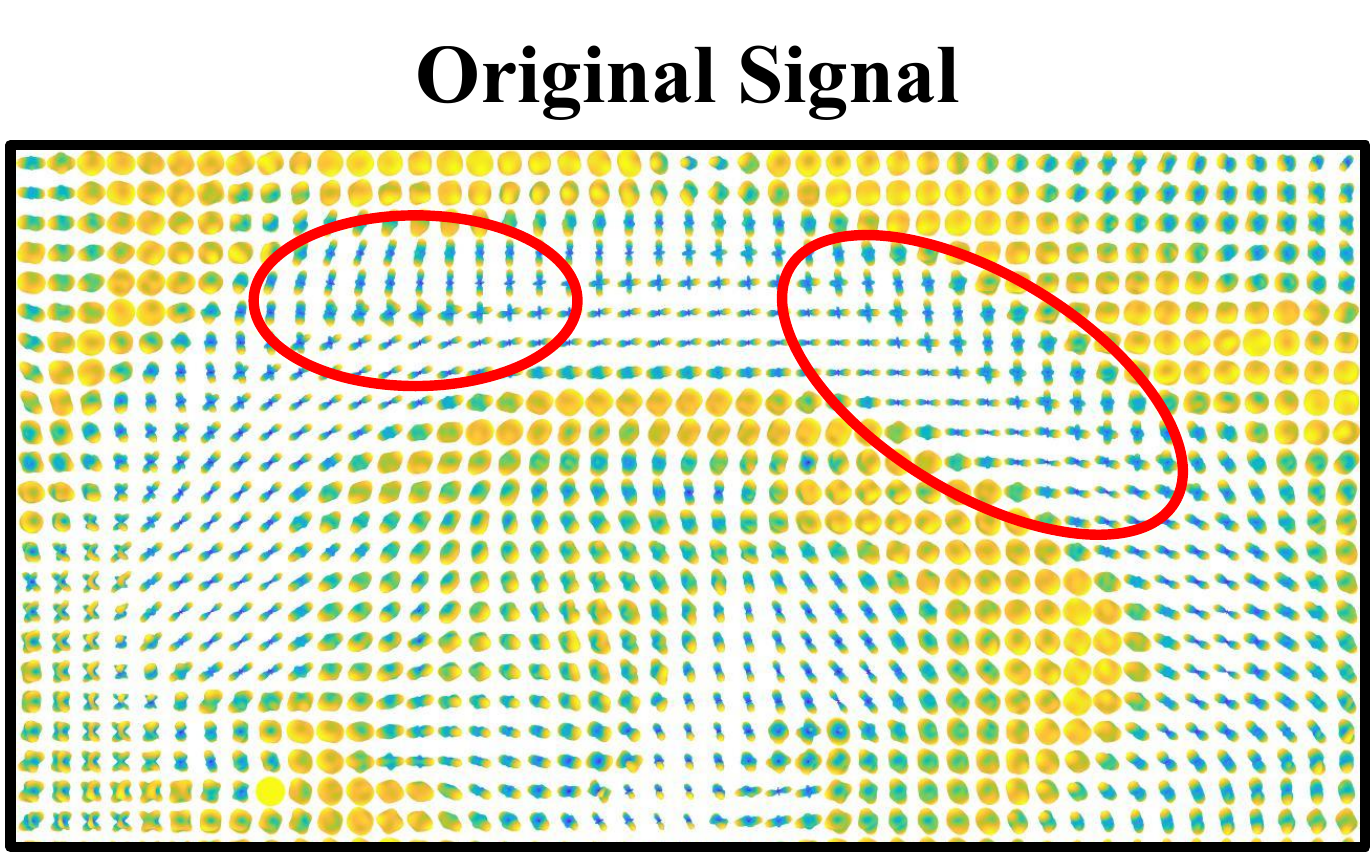}\\
\includegraphics[width=.49\linewidth]{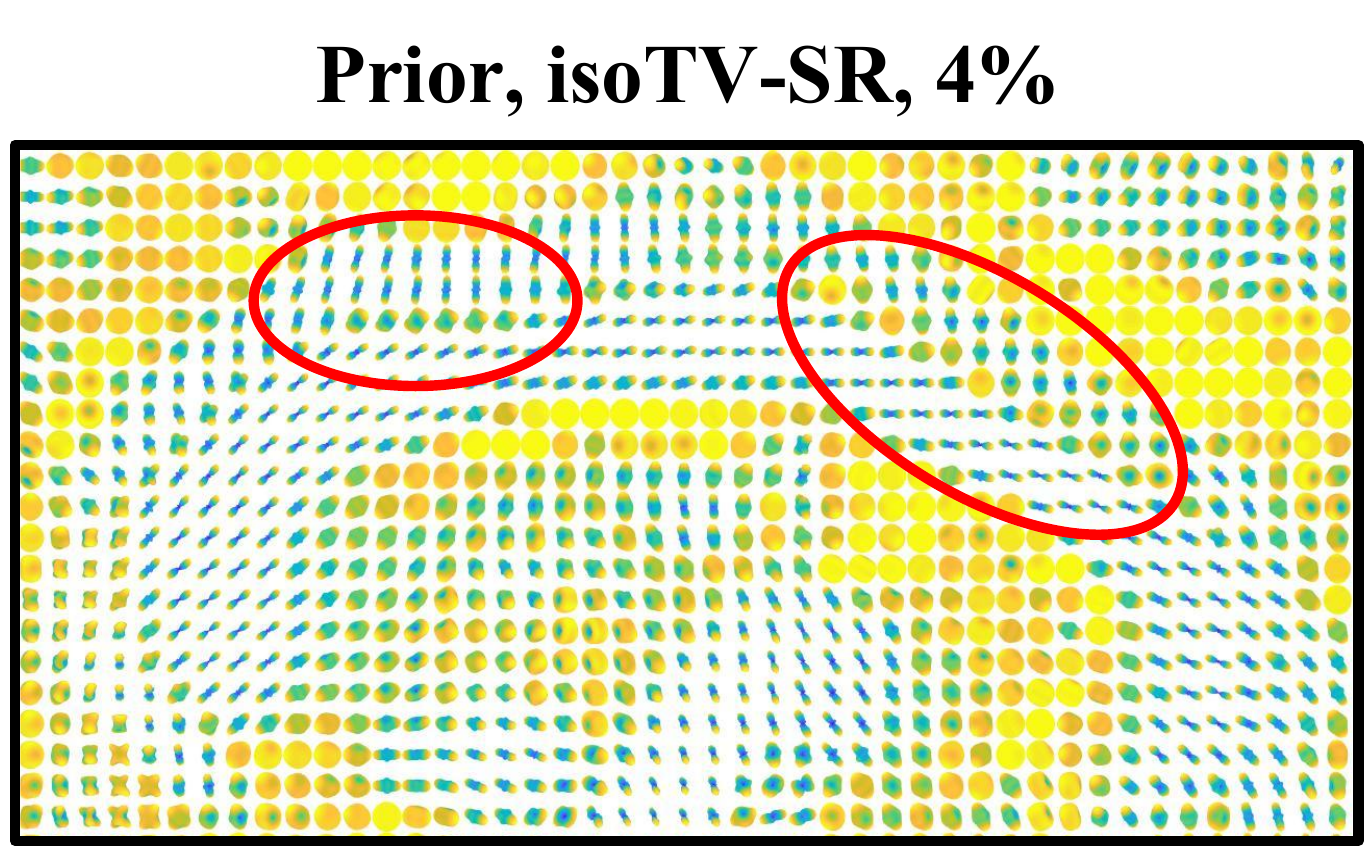}
\includegraphics[width=.49\linewidth]{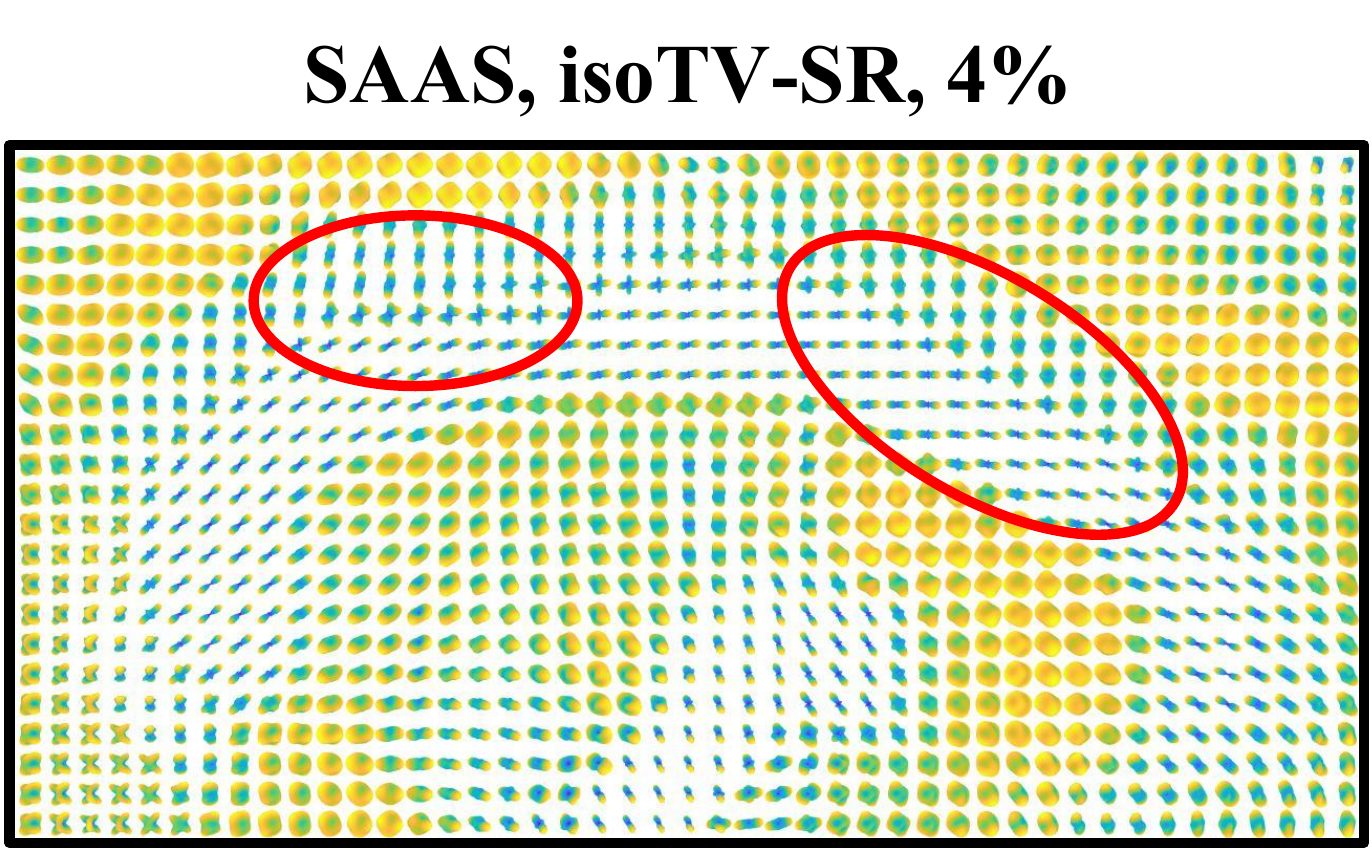}\\
\includegraphics[width=.49\linewidth]{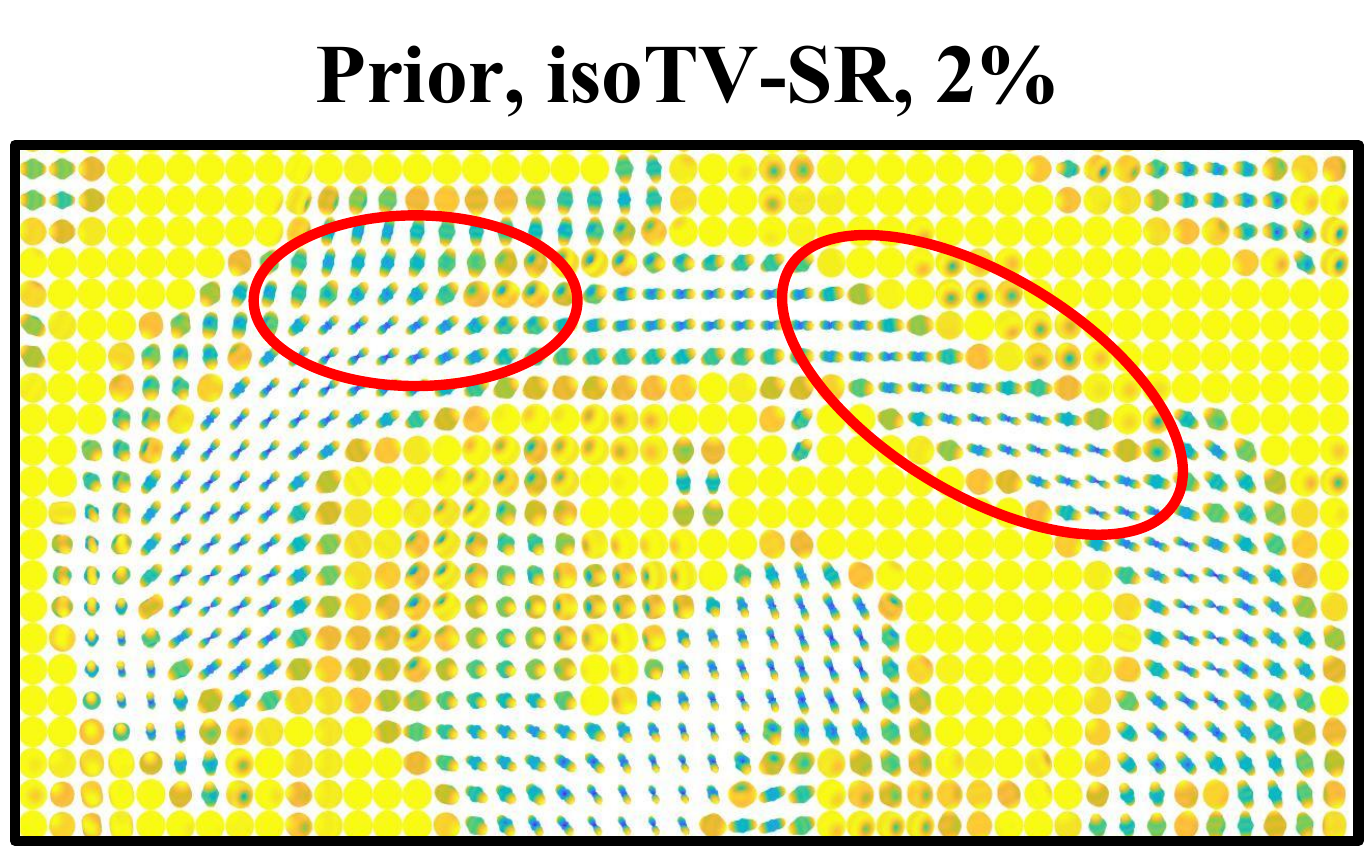}
\includegraphics[width=.49\linewidth]{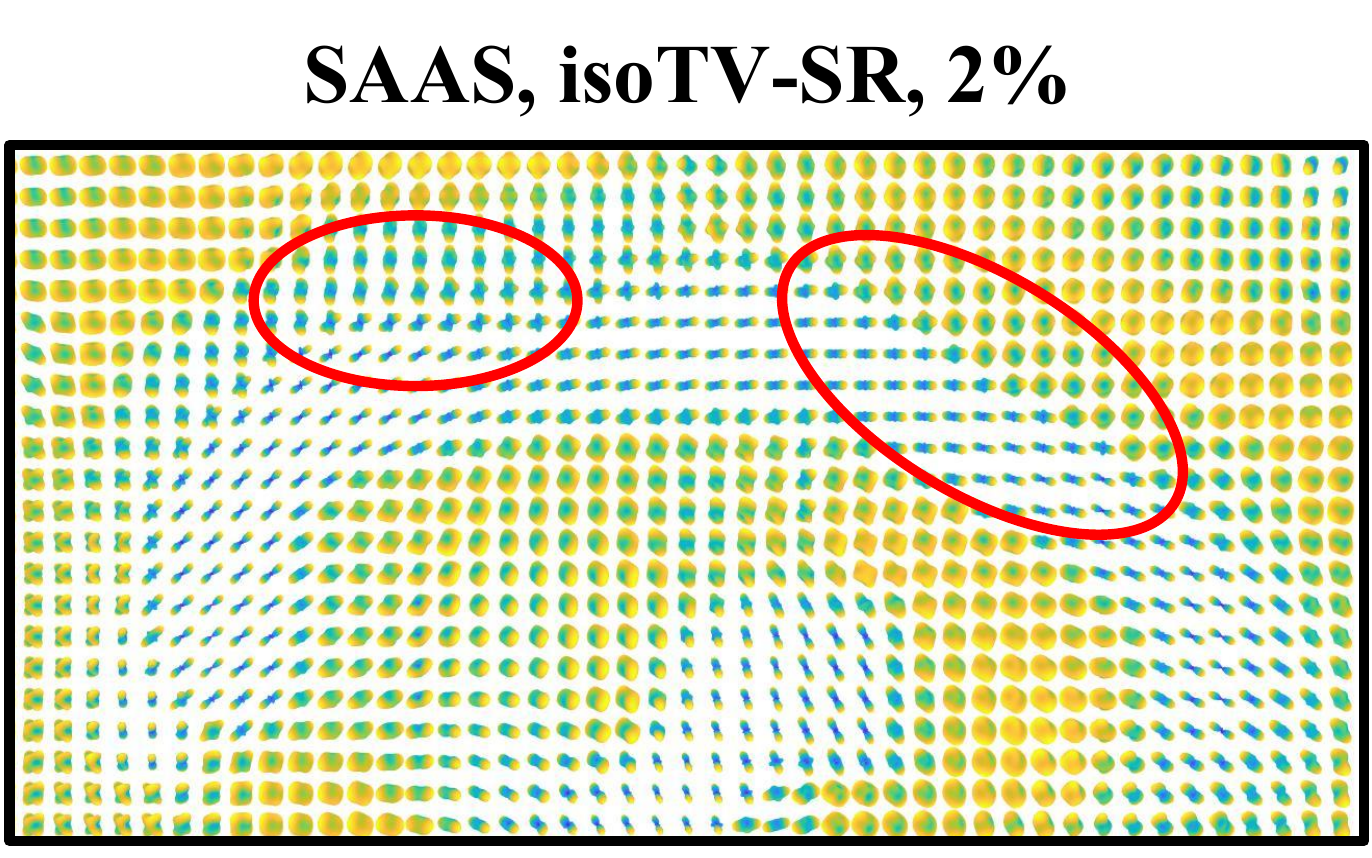}
\caption{Reconstruction of corpus callosum in the sagittal view comparing \eqref{eq:kqCSmat_A} and \eqref{eq:kqCSmat_sep_A} ($k,q$)-CS. Top left: whole brain $b0$ image with ROI. Top right: ODFs in ROI estimated from fully sampled original signal. Middle: ODFs estimated from reconstructed signal with only $4\%$ of the total ($k,q$) measurements, keeping $20\%$ $k$-space samples and $20\%$ $q$-space samples (51 grad dirs). Bottom: repeated with $2\%$ of the total ($k,q$) measurements, keeping only $10\%$ $q$-space samples (25 grad dirs). \eqref{eq:kqCSmat_sep_A} is unable to reconstruct crossing fibers and sets many voxels to zero (yellow) while \eqref{eq:kqCSmat_A} maintains accurate reconstruction at these very low sampling rates.}
\label{fig:kqsubqualreal}
\end{figure}

We next show ($k,q$)-CS results on a real HARDI brain dataset with $G=256$ gradient directions ($b=1500$ s/mm$^2$). For visualization we tested on a 2D $50\times 50$ sagittal slice of the corpus callosum region known for two distinct crossing fiber tract populations in the left-right and anterior-posterior directions.  Fig.~\ref{fig:kqsubqualreal} shows the results of our proposed \eqref{eq:kqCSmat_A} vs. \eqref{eq:kqCSmat_sep_A} first with $20\%$ $k$-space and $20\%$ $q$-space (51 gradient directions) subsampling and then decreased to $20\%$ $k$-space and $10\%$ $q$-space (25 gradient directions) for a total of $4\%$ and $2\%$ subsampling, respectively. We can see that at $4\%$, \eqref{eq:kqCSmat_A} is able reconstruct the crossing ODFs in this region while \eqref{eq:kqCSmat_sep_A} results in isotropic estimations.  As we decrease subsampling further to $2\%$, we notice that \eqref{eq:kqCSmat_sep_A} produces a highly inaccurate reconstruction, setting many voxels to zero (yellow spheres). \eqref{eq:kqCSmat_A} maintains a recognizable structure but begins to lack accuracy of crossing fibers. 

As was the case for phantom data, this real data has been pre-transformed to the spatial domain and so we retrospectively transform the data back to $k$-space using the Fourier transform in order to subsample the data before experiments. Applying our methods directly to raw data acquired in $(k,q)$-space will be included in future work. These results show the limitations of subsampling for state-of-the-art $(k,q)$-CS, and the promise of new levels of subsampling and acceleration using our proposed ($k,q$)-CS model.
\section{Conclusion}
\label{sec:conclusion}
In this work, we have proposed a unified ($k,q$)-CS model for dMRI that naturally exploits sparsity in the joint spatial-angular domain. The main goal of this paper was to demonstrate the performance gains of CS using our joint model compared to state-of-the-art frameworks which combine $k$-CS and $q$-CS in an additive way. We have shown that we can achieve more accurate signal reconstructions with a greater reduction of measurements than state-of-the-art ($k,q$)-CS models, on the order of $2$-$4\%$ of the original data (12-25 gradient directions).  Though we experimented on single-shell HARDI, our proposed framework is applicable to any dMRI acquisition protocol.  In addition, we have derived a novel Kronecker extension of FISTA to efficiently solve this large-scale optimization by exploiting the separability of Kronecker dictionaries.  

To make a concrete comparison of ($k,q$)-CS methods, we chose fixed sparsifying transforms/dictionaries and $(k,q)$ sensing schemes and used a spatial-angular \textit{analysis}-\textit{synthesis} model to match that of state-of-the-art formulations.  In our future work, we will develop joint spatial-angular dictionary learning methods to increase sparsity and optimize ($k,q$) sensing schemes to push the limits of acquisition acceleration.  We will also explore other \textit{analysis}/\textit{synthesis} options and derive precise theoretical guarantees for the proposed $(k,q)$-CS model. Lastly, in future work we will closely investigate the relationship between sampling in $(k,q)$-space as a function of acquisition time.  We hope that the preliminary findings in this work may lead to increased levels of dMRI acceleration for greater practical usability in the future.

\myparagraph{Acknowledgement} This work was supported by JHU start-up funds.

\bibliographystyle{plain}
{\footnotesize
\bibliography{vidal,dti,math,biomedical,sparse}}

\end{document}